\documentclass[letterpaper, 10 pt, conference]{ieeeconf}
\IEEEoverridecommandlockouts

\title{\LARGE \bf
Adaptive Contact-Implicit Model Predictive Control with Online Residual Learning
}

\author{Wei-Cheng Huang*, Alp Aydinoglu*, Wanxin Jin, and Michael Posa
\thanks{*Contributed equally to this work. Wei-Cheng Huang, Alp Aydinoglu and Michael Posa are with the GRASP Laboratory, University of Pennsylvania, USA. Wanxin Jin is with the School for Engineering of Matter, Transport and Energy, Arizona State University, USA.
        {\tt\small \{wchuang, alpayd, posa\}@seas.upenn.edu, wanxin.jin@asu.edu}}%
\thanks{This work was supported by the National Science Foundation under Grants No. FRR-2238480, EFRI-1935294, and Toyota Research Institute.}
}

\usepackage{cite}
\usepackage{amsmath,amssymb,amsfonts}
\usepackage{algorithmic}
\usepackage{graphicx}
\usepackage{textcomp}
\usepackage{xcolor}
\def\BibTeX{{\rm B\kern-.05em{\sc i\kern-.025em b}\kern-.08em
    T\kern-.1667em\lower.7ex\hbox{E}\kern-.125emX}}

\usepackage{algorithm}
\usepackage[hidelinks]{hyperref}

\newtheorem{theorem}{Theorem}

\newtheorem{definition}[theorem]{Definition}

\newcommand{\rev}[1]{\textcolor{black}{#1}}
    
\begin{document}


\maketitle
\thispagestyle{empty}
\pagestyle{empty}

\begin{abstract}
The hybrid nature of multi-contact robotic systems, due to making and breaking contact with the environment, creates significant challenges for high-quality control. Existing model-based methods typically rely on either good prior knowledge of the multi-contact model or require significant offline model tuning effort, thus resulting in low adaptability and robustness. In this paper, we propose a  real-time adaptive multi-contact model predictive control framework, which enables online adaption of the hybrid multi-contact model and continuous improvement of the control performance for contact-rich  tasks. This framework includes an adaption module, which  continuously learns a residual of the hybrid model to minimize the gap between the prior model and reality, and a real-time multi-contact MPC controller. We demonstrated the effectiveness of the framework in synthetic examples, and  applied it on hardware to solve contact-rich manipulation tasks, where a robot uses its end-effector to roll different unknown objects on a table to track given paths. The hardware experiments show that with a rough prior model, the multi-contact MPC controller adapts itself on-the-fly with an adaption rate around $20$ Hz and successfully manipulates previously unknown objects with non-smooth surface geometries. Accompanying media can be found at:
{\color{black}\url{https://sites.google.com/view/adaptive-contact-implicit-mpc/home}}

\end{abstract}


\section{Introduction}

\rev{For in-home or workplace robots to achieve their true potential, assisting humans across a wide range of tasks in a complex and cluttered environments, they must be capable of safely and quickly reacting to that complexity.
One key challenge to achieving high-performance control for multi-contact robotic tasks, particularly dexterous manipulation, lies in the combinatoric complexity of simultaneously sequencing contact locations and selecting continuous control actions. 
Significant progress has been made in this area, with recent methods demonstrating real-time multi-contact (MPC) \cite{tassa2012synthesis, aydinoglu2019contact,cleac2021fast}, but these methods require an accurate multi-contact model, with control performance ultimately limited by model accuracy and complexity.
However, acquiring such a model is difficult in real-world settings, and so performance on these multi-contact robotic tasks, particularly dexterous manipulation, remains limited to laboratory settings and controlled demonstrations.}



\rev{While classical approaches exist to modeling or identifying the continuous dynamics, we observe that multi-contact control is particularly sensitive to accurate predictions of the contact events (e.g. when a robot will make or break contact with an object).
To address this requirement for a model, recent work (e.g. \cite{jin2022task,chen2020optimal}) has started with system identification, }using collected data to learn a model\cite{jin2022learning,pfrommer2021contactnets,qi2023hand,kumar2021rma} to improve the control performance for contact-rich tasks. Despite the success, the offline model learning in those methods requires a significant effort of data collection and training to obtain an effective model, limiting the flexibility of these methods when deployed in practice. To address those limitations, in this work, we take a perspective from the classic adaptive control paradigm \cite{sastry1990adaptive} but specifically address multi-contact systems: we present an adaptive multi-contact MPC controller such that it can adjust the multi-contact hybrid model in real time to account for variations of unknown objects/environments and improve its control performance for multi-contact tasks.

\begin{figure}[t!]
	\centering
        \includegraphics[width=0.49\columnwidth]{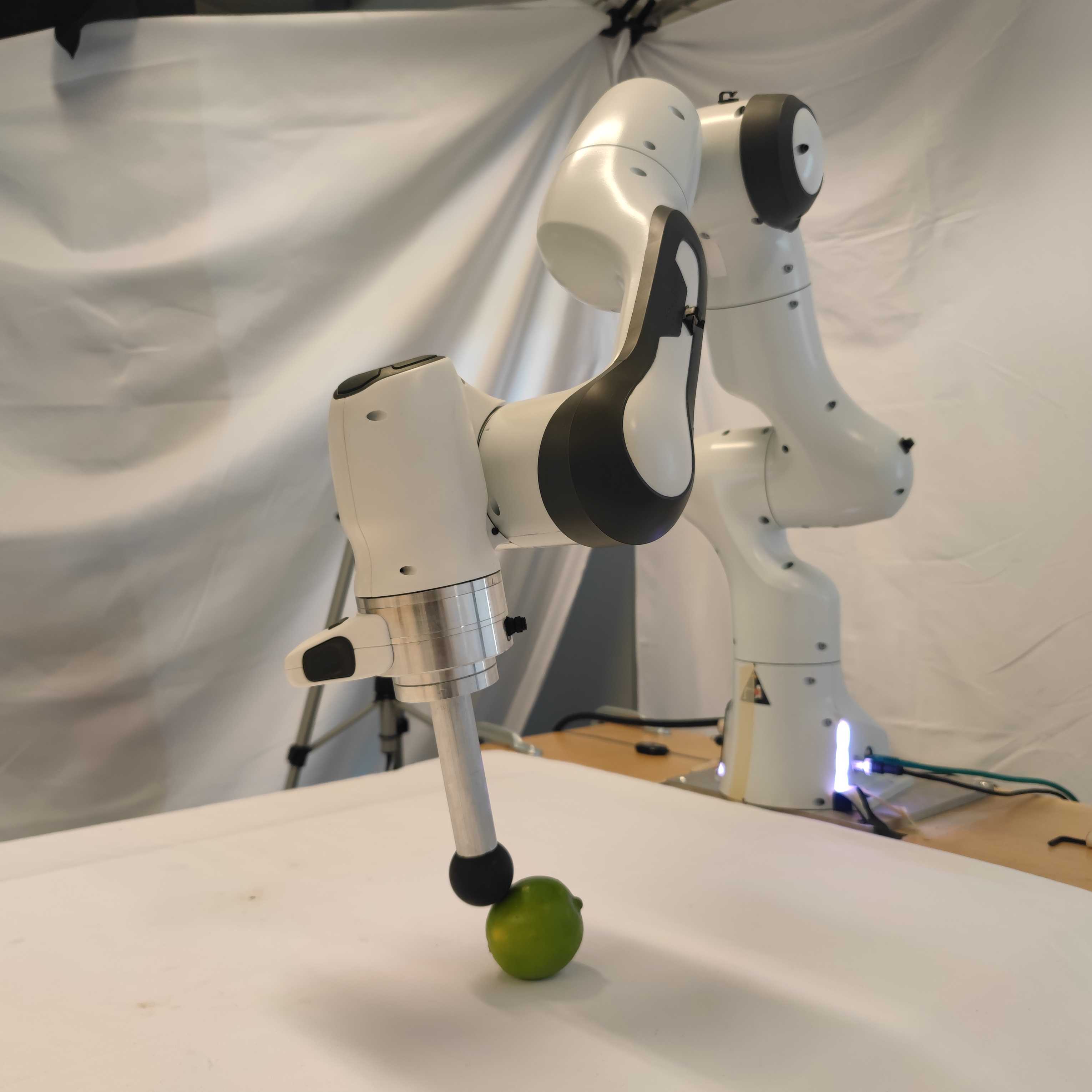}
	\includegraphics[width=0.49\columnwidth]{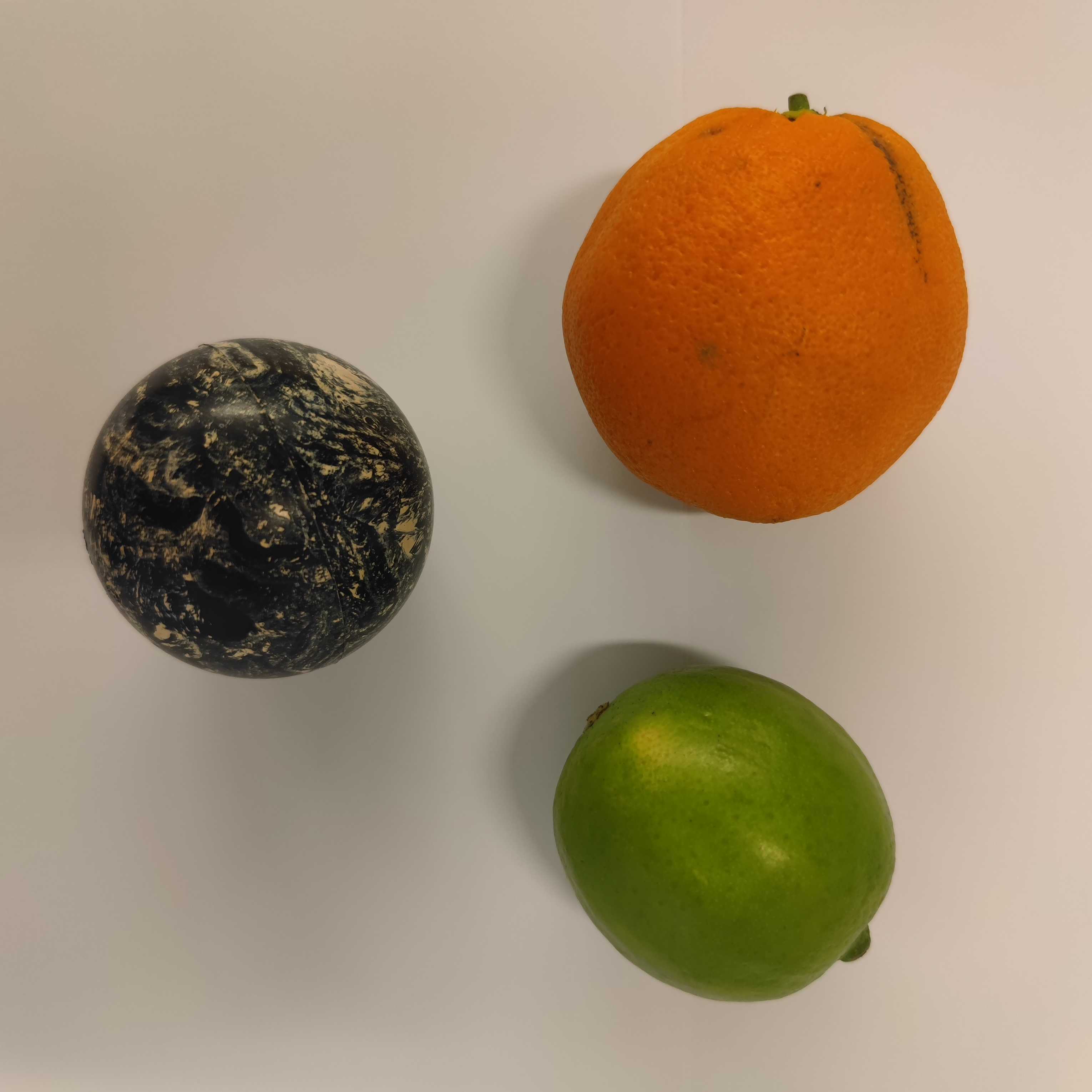}
	\caption{Given an initial guess that the object is a rigid sphere, the controller adapts its model of the governing contact dynamics to roll and push real fruits (orange, lime) with a Franka Emika Panda arm, tracking a desired motion.}
	\label{franka}
 \vspace{-10pt}
\end{figure}

The presented adaptive framework continuously adjusts the hybrid model, using real-time in-stream data; meanwhile, the MPC controller uses the most recent model for high-quality control on multi-contact tasks. 
Compared to the existing work, we emphasize the two key contributions below.

(i) We present an adaptation law for multi-contact systems that continuously minimizes the gap between the prior model and reality by learning a hybrid residual model from in-stream data. It is inspired by our \emph{offline} learning method \cite{jin2022learning} but unlike the previous work it focuses on \emph{online} learning and utilizes physics-based prior models to speed up the process. Furthermore, the hybrid model adaption module is integrated into our fast multi-contact MPC controller \cite{aydinoglu2023consensus} which allows real-time adaptive contact-implicit control on robots with on-the-fly performance improvement for contact-rich tasks.

(ii) We validate the proposed approach with two different hardware experiments, accomplishing challenging robot arm manipulation tasks. Our method adapts itself on-the-fly with an adaption rate around $20$ Hz and can consistently be employed to manipulate objects with uneven or irregular surface geometries. The findings also indicate that our approach can succeed in tasks that purely model-based methods fail. 

\section{Related Work}

\subsubsection{Learning Multi-Contact Dynamics Models}
The hybrid nature of multi-contact dynamics poses challenges for gradient-based learning. One line of work, termed differentiable simulation, focuses on smoothing hybrid mode boundaries, although this can lead to approximation errors \cite{geilinger2020add,heiden2020augmenting,howell2022dojo}. Recently, however, results have shown that conventional neural networks are limited in capturing the multi-modality and high-stiffness of multi-contact systems \cite{parmar2021fundamental,pfrommer2021contactnets,bianchini2022generalization}.
To address this, researchers \cite{pfrommer2021contactnets,jin2022learning,bianchini2022generalization} explicitly exploit the hybrid structure (complementarity formulation) of contact dynamics to develop learning algorithms, achieving state-of-the-art performance. In our adaptive contact-implicit MPC framework, the adaption module focuses on learning a hybrid residual model on top of prior dynamics,  which is based on our previous work \cite{jin2022learning}. The benefit of  \cite{jin2022learning} is that it simultaneously performs the mode partitioning and linear regression by proposing a novel training loss. The quadratic formulation in the loss enables fast updates using in-stream data, which can run up to $20$ Hz.

\subsubsection{Fast Multi-Contact Model Predictive Control}
To handle the combinatoric complexity of choosing hybrid contact modes, previous work \cite{sleiman2021unified,mastalli2020crocoddyl} use pre-defined sequence of modes to achieve real-time multi-contact control
on legged locomotion \cite{winkler2018gait} and manipulation \cite{hogan2020reactive}. To achieve contact-implicit control, the work \cite{cleac2021fast} proposed to relax contact mode boundaries with smoothing approximation. Concurrently, our work \cite{aydinoglu2023consensus} introduced an approach that preserves the hybrid structures while decoupling the combinatorial complexity from planning depth. However, rather than operating with an accessible ground-truth multi-contact model, we incorporate a hybrid model adaptation module into the MPC controller. This module operates continuously, updating the hybrid model with data collected from real-world interactions. Consequently, the MPC controller can consistently improve its performance for multi-contact tasks.

\subsubsection{Task-Oriented Multi-Contact Model Learning}

A relevant body of recent research is task-driven multi-contact model learning, which aims to find a model that can be used to (derive a policy and) accomplish given tasks. This line of work shares the same idea as deep model-based reinforcement learning \cite{nagabandi2020deep}, although the latter typically suffers from huge data demand. The most recent work focuses on learning a task-driven computationally affordable models, which have been shown requiring a small amount of data to successfully solve deleterious manipulation \cite{jin2022task} and bipedal locomotion \cite{chen2020optimal}.
It is important to note that the above existing methods predominantly rely on offline model learning. This means that the model undergoes episodic updates after accumulating a buffer of historical policy data. In contrast, the approach presented in this paper emphasizes continuous and online updates to the hybrid residual model. This innovative method facilitates real-time improvements in the multi-contact hybrid model, consequently improving performance during the deployment of the MPC policy. Additionally, this adaptation framework holds the potential to reduce data consumption when compared to offline model learning methodologies.

\subsubsection{Adaptive MPC} The most relevant theme of this work is adaptive control, which focuses on control of uncertain systems through real-time model adaptation and learning \cite{annaswamy2021historical} and has long been an ongoing research direction  \cite{becker1985adaptive}. Researchers have developed robust (adaptive) MPC methods \cite{tanaskovic2014adaptive, fukushima2007adaptive, weiss2014robust, desaraju2017experience}, which are successfully applied to  electric vehicles \cite{wang2018efficient}, climate control \cite{ma2012fast}, quadrotors \cite{pereida2018adaptive}. Similarly, there are multiple recent methods that perform adaptive MPC under unknown noise distributions \cite{stamouli2022adaptive} or incorporate components from adaptive control into learning-based MPC \cite{chee2023enhancing}. Also, the recent work \cite{guzman2022adaptive} presents an adaptive MPC variant that automatically estimates control and model parameters by leveraging ideas from Bayesian optimization performing manipulation tasks. Unlike our approach, which focuses on real-time adaptive MPC for systems that make and break contact, the authors focus on reaching tasks, avoiding any contact interaction.

\section{Background}


\subsection{Multi-Contact Dynamics}

A common model for the multi-contact dynamics of a rigid manipulator is
\begin{align}
	\label{eq:manipulator_equation}
	& M(q) \dot{v} + C(q,v) = Bu + J(q)^T \lambda, \\
 \label{eq:manipulator_complementarity_constraints}
	& 0 \leq \lambda \perp \psi (q,v, u, \lambda) \geq 0.
\end{align}
where $q$ and $v$ are vectors of the generalized positions and velocities, respectively, and $\lambda$ represents the contact forces. 
The complementarity constraint \eqref{eq:manipulator_complementarity_constraints} defines the hybrid rellationship between the contact forces and the generalized positions, velocities, and inputs (see \cite{brogliato1999nonsmooth, stewart2000implicit, anitescu2006optimization} for full details).
We note that one might equivalently explicitly represent \eqref{eq:manipulator_equation} as $\dot{x} = f(x, \lambda, u)$ with $x = \begin{bmatrix}
    q^T,  v^T 
\end{bmatrix}^T $.

\subsection{Linear Complementarity Problem}
\label{subsec:lcp}
While contact dynamics are represented via nonlinear complementarity constraints \eqref{eq:manipulator_complementarity_constraints} ($\psi$ nonlinear in $\lambda$), local models frequently use \textit{linear} complementarity problems (LCPs) as a means to depict contact forces \cite{stewart2000implicit, anitescu2006optimization,cottle2009linear}.
\begin{definition} \label{def: LCS_definition}
	Given a vector $q \in \mathbb{R}^m$, and a matrix ${F \in \mathbb{R}^{m \times m}}$, the $\text{LCP}(q,F)$  describes the following  program:
	\begin{alignat}{2}
		\label{LCS_definiton}
		\notag & \underset{}{\text{find}} && \lambda \in \mathbb{R}^m \\
		\notag & \text{subject to}  \quad && y = F \lambda + q,\\
		\notag & \quad && 0 \leq \lambda \perp y \geq 0.
	\end{alignat}
\end{definition}

\subsection{Linear Complementarity Systems}
\label{subsec:lcs}
Linear complementarity systems (LCS) embed LCPs into dynamical systems, which we employ as localized models for multi-contact systems \cite{aydinoglu2020stabilization, aydinoglu2023consensus, heemels2000linear}. 
\begin{definition}
    An LCS describes the trajectories $( x_k )_{k \in \mathbb{N}_{0}}$ and $( \lambda_k )_{k \in \mathbb{N}_{0}}$ for an input sequence $( u_k )_{k \in \mathbb{N}_{0}}$ such that
	\begin{equation}
	\label{eq:LCS}
	\begin{aligned}
	& x_{k+1} = A x_k + B u_k + D\lambda_k + d,\\
	& 0 \leq \lambda_k \perp Ex_k +  F \lambda_k + H u_k + c \geq 0,
	\end{aligned}
	\end{equation}
	for a given $x_0$ where $x_k \in \mathbb{R}^{n_x}$, $\lambda_k \in \mathbb{R}^{n_{\lambda}}$, $u_k \in \mathbb{R}^{n_u}$.
\end{definition}

Given $x_k$ and $u_k$, we can determine the associated complementary variable $\lambda_k$ by solving the linear complementarity problem $\text{LCP}(E x_k + H u_k + c, F)$ (Definition \ref{def: LCS_definition}).
Moving forward, we describe the set of matrices in the LCS model \eqref{eq:LCS} as $\theta = \{ A, B, D, d, E, F, H, c \}$.
\rev{If the elements depend on a given nominal (denoted using $*$) state-input pair ($x^*, u^*$), we represent them as $A^* = A(x^*, u^*)$ and the set of such matrices is denoted as $\theta^*$.} 
With a slight abuse of notation, we denote the state-contact force pair an LCS generates as $(x_{k+1}, \lambda_{k}) = \mathcal{L}(x_k, u_k, \theta^*)$.

\section{Problem Formulation}

In this work, we are interested in solving the following MPC problem at real-time rates:
\begin{equation}
\label{eq:MPC_original}
\begin{aligned}
\min_{x_k, \lambda_k, u_k} \quad & \sum_{k=0}^{N-1} (x_k^T Q_k x_k + u_k^T R_k u_k) + x_N^T Q_N x_N \\
\textrm{s.t.} \quad &x_{k+1} = A^* x_k + B^* u_k + D^* \lambda_k + d^* + r_{\text{dyn}}, \\
& 0 \leq \lambda_k \perp E^* x_k + F^* \lambda_k + H^* u_k + c^* + r_{\text{comp}} \geq 0,
\end{aligned}
\end{equation}
where $N$ is the planning horizon, $Q_k, Q_N$ are positive semidefinite matrices, $R_k$ are positive definite matrices, $r_{\text{dyn}}$ and $r_{\text{comp}}$ represents the residuals for the dynamics and complementarity terms respectively (these could be time or state dependent). If one has perfect knowledge of the system matrices, $\theta^*$, then $r_{\text{dyn}} = r_{\text{comp}} = 0$. In our previous work, we presented a framework for the setting where we relied on high model accuracy, e.g. $r_{\text{dyn}}~\approx~r_{\text{comp}} \approx 0$. 
While \eqref{eq:MPC_original} is non-convex, we leverage a recent algorithm called Consensus Complementarity Control (C3) to achieve real-time, albeit subptimal, solutions (see \cite{aydinoglu2023consensus} for details on C3).

In this work, we focus on the setting where we have access to an imperfect physics-based model, e.g. $\theta^*$ is inaccurate, and where the error is significant enough to prevent task completion. Our aim is to collect data online and adaptively learn a residual $(r_{\text{dyn}},r_{\text{comp}} )$  while simultaneously solving the MPC problem, both at real-time rates. Our goal is that, over time, our adaptive model accurately captures the behavior of the true dynamics, and that $u_0$ well-approximates the solution of \eqref{eq:MPC_original} with true dynamics. We highlight that we want to achieve both goals at real-time rates. 

\section{Physics-Based LCS}
\label{sec:anitescu}

This section describes the process of converting Anitescu's approach for simulating contact dynamics \cite{anitescu2006optimization} into an LCS approximation around a given nominal state-input pair $(x^*, u^*) \to \theta^*$.
We choose Anitescu's formulation over other approaches, such as the Stewart-Trinkle formulation \cite{stewart2000rigid}, because it is a convex contact model (i.e. $F^*$ as in Section \ref{subsec:lcs} is positive semi-definite for any $(x^*, u^*)$) and our proposed learning algorithm (discussed in Section \ref{sec:residual}) relies on this assumption.
Next, we describe our physics based model which is a discrete-time approximation of \eqref{eq:manipulator_equation}, \eqref{eq:manipulator_complementarity_constraints}:
\begin{equation}
\label{eq:Anitescu_dynamics_nonlinear}
\begin{aligned}
q_{k+1} = q_k &+ \Delta t v_{k+1}, \\
v_{k+1} = v_k &+ M^{-1} \bigg( \Delta t B u_k - \Delta t C(q_k,v_k)  + J_c(q_k)^T \lambda_k \bigg),
\end{aligned}
\end{equation}
with the complementarity constraints:
\begin{equation}
    \label{eq:Anitescu_constraints_nonlinear}
    0 \leq \lambda_k \perp \frac{E_t^T}{\Delta t} \bigg( \phi + J_n (q_k) (q_{k+1} - q_k) \bigg) + \mu J_t(q_k) v_{k+1} \geq 0.
\end{equation}
Here $\phi$ represents the distance between rigid body pairs, $J_n$, $J_t$ are contact Jacobians for normal and tangential directions, the contact Jacobian is defined as $J_c = E_t^T J_n + \mu J_t$, $\mu$ represents the coefficient of friction and $ E_t = \mathbf{blkdiag}(e, \ldots, e)$ with $e =[1, \ldots,  1] \in \mathbb{R}^{1 \times n_e}$ (where $n_e$ represents the number of edges of the polyhedral approximation of the friction cone).

Given state $(x^*)^T = [(q^*)^T, (v^*)^T]$ and input $u^*$, we can approximate \eqref{eq:Anitescu_dynamics_nonlinear} as:
\begin{equation}
	\label{eq:Anitescu_linear_dynamics}
	\begin{aligned}
	& q_{k+1} = q_k + \Delta t v_{k+1}, \\
	& v_{k+1} = v_k + \Delta t J_f \begin{bmatrix} q_k^T & v_k^T & u_k^T \end{bmatrix}^T + D \lambda_k + d_v,
	\end{aligned}
\end{equation}
where $J_f = J_f(q^*, v^*, u^*)$ is the Jacobian of $f(q,v,u) = M^{-1}(q) B u - M^{-1}(q) C(q,v)$ evaluated at $(q^*, v^*, u^*)$, $d_v = f(q^*, v^*, u^*) - J_f \begin{bmatrix} (q^*)^T & (v^*)^T & (u^*)^T \end{bmatrix}^T$ is a constant vector and $D = M^{-1}(q^*) J_c(q^*)^T$. Similarly the equation \eqref{eq:Anitescu_constraints_nonlinear} can be approximated as:
\begin{align}
	\label{eq:Anitescu_linear_constraints}
     0 \leq \lambda_k \perp \frac{1}{\Delta t}E_t^T \bigg( \phi(q^*) & + J_n(q^*) q_k - J_n(q^*) q^* \bigg) \\
     & + J_c(q^*) v_{k+1}  + \epsilon_c \lambda_k \geq 0. \notag
\end{align}
We note that because simulation considers single-step predictions, and MPC requires multi-step predictions, \eqref{eq:Anitescu_linear_dynamics}-\eqref{eq:Anitescu_linear_constraints} differs slightly from the LCS presented in previous work \cite{anitescu2006optimization}. These equations 
can be written in the LCS format \eqref{eq:LCS} and represented as $\mathcal{L}(\theta^*)$. 
We have also introduced a regularizing term $\epsilon_c \lambda_k$ in \eqref{eq:Anitescu_linear_constraints} to ensure that $F^*$ is positive definite as Anitescu's formulation produces a positive semi-definite $F^*$.

\section{Adaptive MPC with Residual Learning}
\label{sec:residual}

\begin{algorithm}[t!]
	\caption{}
	\begin{algorithmic}[1]
		\REQUIRE $r_\text{comp}, \mathcal{B}, \epsilon, \gamma, \xi$
		 \STATE Compute $\theta^*$ for each data point and construct $\mathcal{B}^A$
        \STATE Compute the gradient $\nabla_{r_\text{comp}} L_\epsilon(\mathcal{B}^A, r_\text{comp})$
        \STATE Update the residual $r_\text{comp}$ via Adam \cite{kingma2014adam}
	\RETURN $r_\text{comp}$ (updated residual parameter)
	\end{algorithmic} 
	\label{learning_algo}
\end{algorithm}

\begin{figure}[t!]
	\centering
	\includegraphics[width=1\columnwidth]{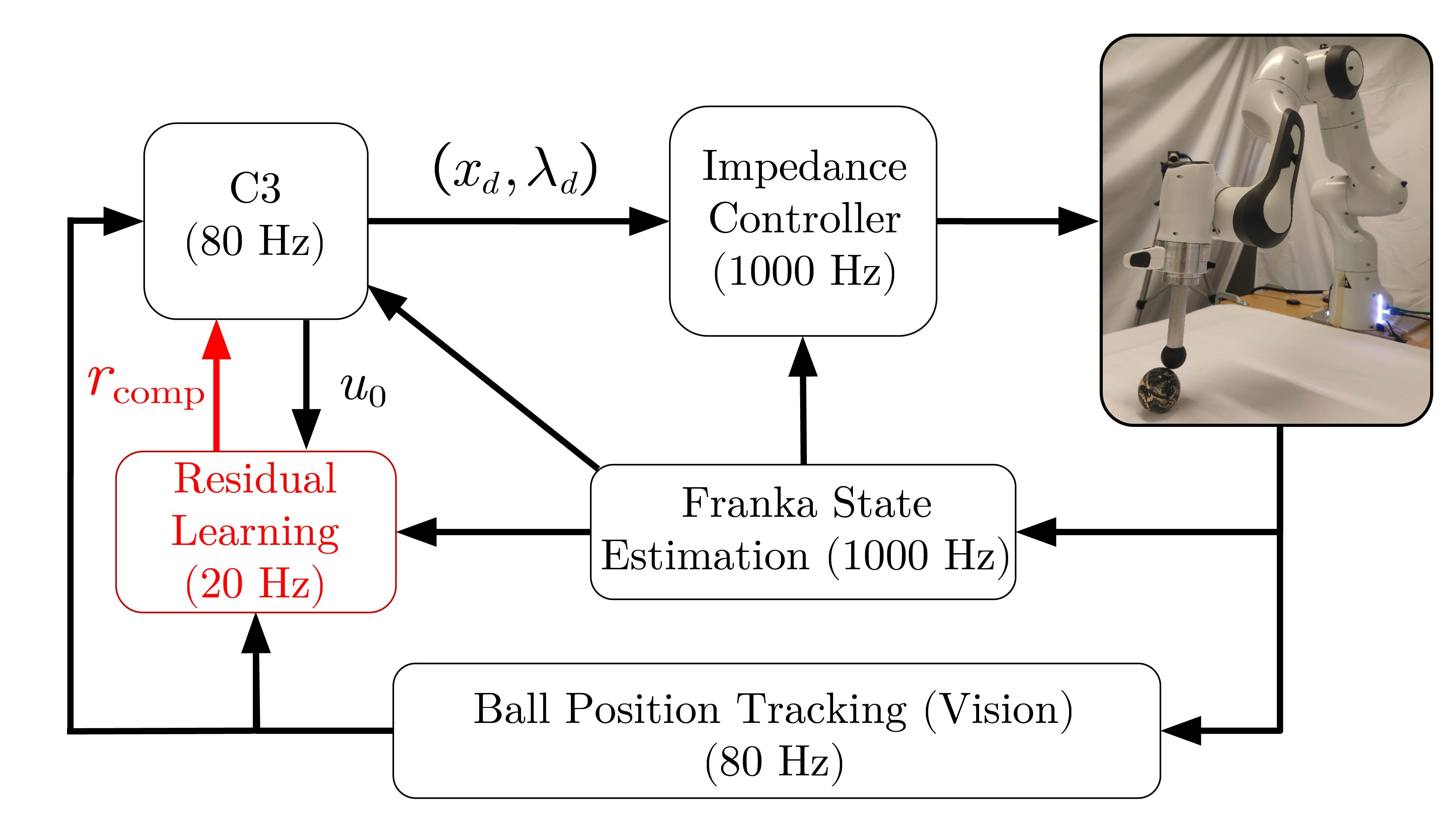}
	\caption{Key elements of the adaptive MPC framework. 
    Given proprioception and visual data, our method learns a residual multi-contact model at $20$ Hz, which we use for real-time control.
    }
	\label{C3_high_level_fig_learning}
 \vspace{-10pt}
\end{figure}

In this section, we describe our adaptive MPC framework shown in Figure \ref{C3_high_level_fig_learning}. First, we describe the residual learning module (shown in red) in detail. Then we give details about the interaction of the residual learning module with C3.


Standard residual learning  \cite{koller2018learning} focuses on models:
\begin{equation}
    \label{eq:residual_general}
    x_{k+1} =  f(x_k, u_k) + r_\text{dyn}(x_k, u_k),
\end{equation}
where the prior model and the learned residual are combined in an additive manner.
If the prior $f$ is an LCS (as in Section \ref{sec:anitescu}), \eqref{eq:residual_general} is equivalent to:
\begin{equation}
\label{eq:LCS_res_upper}
\begin{aligned}
 x_{k+1} & = A^* x_k + B^* u_k + D^* \lambda_k + d^* + r_\text{dyn}(x_k, u_k),\\
& 0 \leq \lambda_k \perp E^* x_k +  F^* \lambda_k + H^* u_k + c^* \geq 0,
\end{aligned}
\end{equation}
and one can use state-of-the-art residual learning frameworks \cite{kulathunga2023residual, torrente2021data} to learn $r_\text{dyn}$.
This form of adaptation is well studied, and not the focus of this paper. Furthermore, because \eqref{eq:LCS_res_upper} does not adapt the hybrid structure encoded in the complementarity constraints, it is doomed to be data inefficient and will struggle to capture the effects of contact \cite{parmar2021fundamental, bianchini2022generalization}. Unlike in prior methods, our focus is on learning both the hybrid boundary and the contact dynamics:
\begin{equation}
\begin{aligned}
 x_{k+1}  = A^* x_k & + B^* u_k + D^* \lambda_k + d^*,\\
 0 \leq \lambda_k   \perp E^* x_k &+ F^* \lambda_k + H^* u_k + c^* \\
  &  + r_\text{comp}(x_k,u_k,\lambda_k)  \geq 0,
\end{aligned}
\end{equation}
where $r_\text{comp}$ represents the model error in contact equation, e.g. \eqref{eq:Anitescu_linear_constraints} and this effect implicitly appears in the dynamics equation.
\rev{More generally, one might try to simultaneously identify residuals for both the continuous and discrete components, e.g. $r_\text{dyn}$ and $r_\text{comp}$.
While disambiguation of the two can be done within this framework \cite{Bianchini2023}, it is difficult in this minimal data regime. Furthermore, we believe that accurate identification of contact events is key to achieving dynamic contact-rich tasks, and thus focusing on the contact residual is critical.} This residual, denoted as $r_\text{comp}$, possesses the ability to capture inaccuracies such as those in normal distance, tangential friction directions and coefficients of friction.
Hence, we consider:
\begin{equation}
\label{eq:LCS_residual}
\begin{aligned}
& x_{k+1}  = A^* x_k + B^* u_k + D^* \lambda_k + d^*,\\
 0 \leq \lambda_k & \perp E^* x_k + F^* \lambda_k + H^* u_k + c^* + r_\text{comp} \geq 0,
\end{aligned}
\end{equation}
where we learn the vector $r_\text{comp}$ adaptively at real-time rates to compensate the error. 
We note that, as in traditional residual learning, our learned residual term $r_\text{comp}$ might be time-varying, and thus also capture state-dependent variations in the complementarity constraints.
Equation \eqref{eq:LCS_residual} consists of physics-based model parameters $\theta^*$, derived (\rev{Section \ref{sec:anitescu}}) from our \emph{prior} knowledge of the system and also the residual vector $r_\text{comp}$ that is learned by collecting data during the experiment. We denote \eqref{eq:LCS_residual} as $\mathcal{L}(x_k, u_k, \theta^*, r_\text{comp})$ that combines the prior model with the learned residual.

\begin{figure*}[t!]
	\centering
	\includegraphics[width=2\columnwidth]{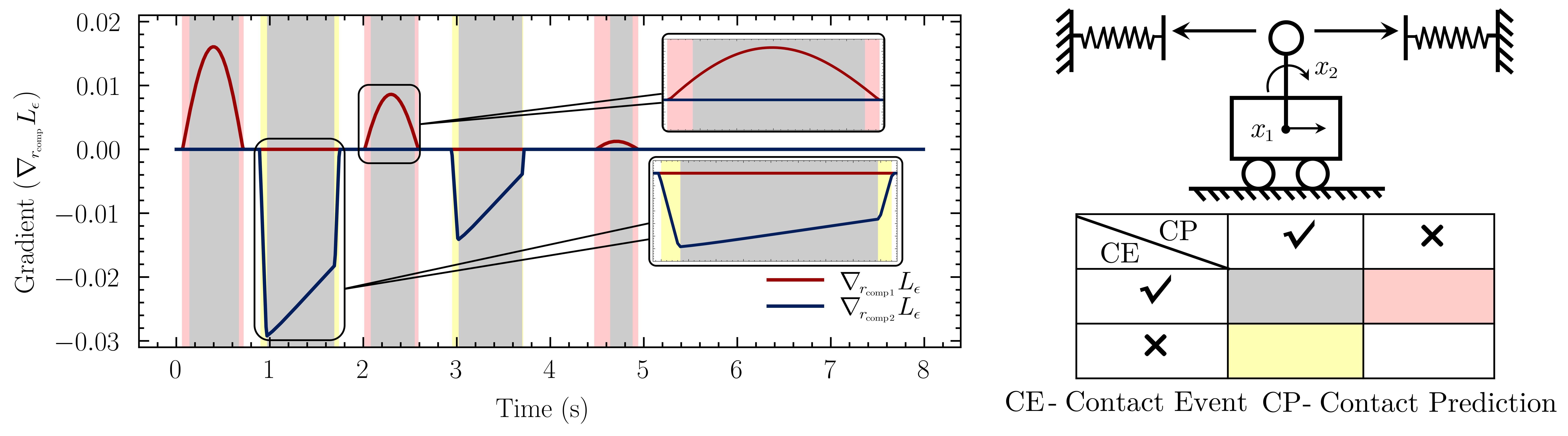}
	\caption{A ``contact event" refers to a situation where actual physical contact is occurring, while ``contact prediction" pertains to instances where the model anticipates contact, potentially inaccurately.
    Our method can produce meaningful gradients, even when there is no actual contact event (yellow region). The only scenario in which a zero gradient is produced is when the model and data both agree that there is no contact (white region). 
 }
	\label{grad}
\end{figure*}

Once an experiment starts, we collect data of the form $\mathcal{B} = \{ x_{k+1}, x_k, u_k \}_{k=n_{ct}}^{n_{ct} + n_b}$ into our buffer where $n_{ct}$ represents the current time-step and $n_b$ is the number of data points stored. 
For each $(x_k, u_k)$, we also calculate the corresponding LCS matrices $\theta_k^*$ (consists of $A^*_k, B^*_k, \ldots$)\footnote{As in Section \ref{subsec:lcs}, $A_k^* = A(x^*, u^*)$ where $(x^*,u^*) = (x_k,u_k)$.} via the method described in Section \ref{sec:anitescu} and define the buffer appended with those matrices as $\mathcal{B}^A = \{ x_{k+1}, x_k, u_k, \theta_k^* \}_{k=n_{ct}}^{n_{ct} + n_b}$. 
Next, we introduce a variant of the implicit loss function that was defined in \cite{jin2022learning}:
\begin{equation*}
    L_\epsilon (\mathcal{B}^A, r_\text{comp} ) = \sum_{k = n_{ct}}^{n_{ct} + n_b} l_\epsilon ( x_{k+1}, x_k, u_k, \theta^*_k, r_\text{comp} ),
\end{equation*}
where the function $l_\epsilon (x_{k+1}, x_k, u_k, \theta^*_k, r_\text{comp} )$ is defined as
\begin{align*}
    l_\epsilon (\cdot) &= \min_{\lambda_k \geq 0, \eta_k \geq 0} \frac{1}{2} (D^*_k \lambda_k + z)^T Q_d (D^*_k \lambda_k + z) \\
    & + \frac{1}{\epsilon} \big( \lambda_k^T \eta_k + \frac{1}{2 \gamma} \big| \big| q +  F^*_k \lambda_k  + r_\text{comp} - \eta_k \big| \big|^2 \big), \notag
\end{align*}
where $z = A_k^* x_k + B_k^* u_k + d_k^* - x_{k+1}$ and $q = E^*_k x_k + H^*_k u_k + c^*_k$. Here, $\gamma$ is a constant such that $0 < \gamma < \sigma_{\min}\big( (F_k^*)^T + F_k^* \big)$ for all $k$ where $\sigma_{\min}(\cdot)$ denotes the smallest singular value. It is important to note that based on our specific formulation using Anitescu's method, all $F_k^*$ are positive definite and we can always find a $\gamma$ that satisfies the given equality. Similarly, $\epsilon$ is a constant such that $\epsilon > 0$. Under these conditions, we can calculate the gradient of the loss function with respect to the residual parameter $r_{\text{comp}}$, i.e. $\nabla_{r_{\text{comp}}} L_\epsilon(\cdot)$ following our previous work \cite{jin2022learning}. This approach requires solving a single quadratic program per data point in the batch (hence is relatively fast). After gradient calculations, residual parameters are updated with a simple gradient step (via learning rate $\xi > 0$). Following this discussion, Algorithm \ref{learning_algo} summarizes how the proposed learning method works.

Both our residual learning module and C3 run at real-time rates (Figure \ref{C3_high_level_fig_learning}). The MPC algorithm uses the latest residual value $r_\text{comp}$, as well as the current state $x^*$ and computes the optimal input $u_0 = \text{C3}(x^*, \mathcal{L}(x^*, u^*, \theta^*, r_\text{comp}))$.
Then, the desired next state-contact force pair is computed as $(x_d, \lambda_d) = \mathcal{L}(x^*, u_0, \theta^*, r_\text{comp})$. An impedance controller is used to track the desired values \cite{khatib1987unified, hogan1985impedance}.


\section{Examples}

\subsection{Synthetic Example: Cart-pole with Soft Walls}
\begin{figure}[b!]
\vspace{-10pt}
	\centering
	\includegraphics[width=1\columnwidth]{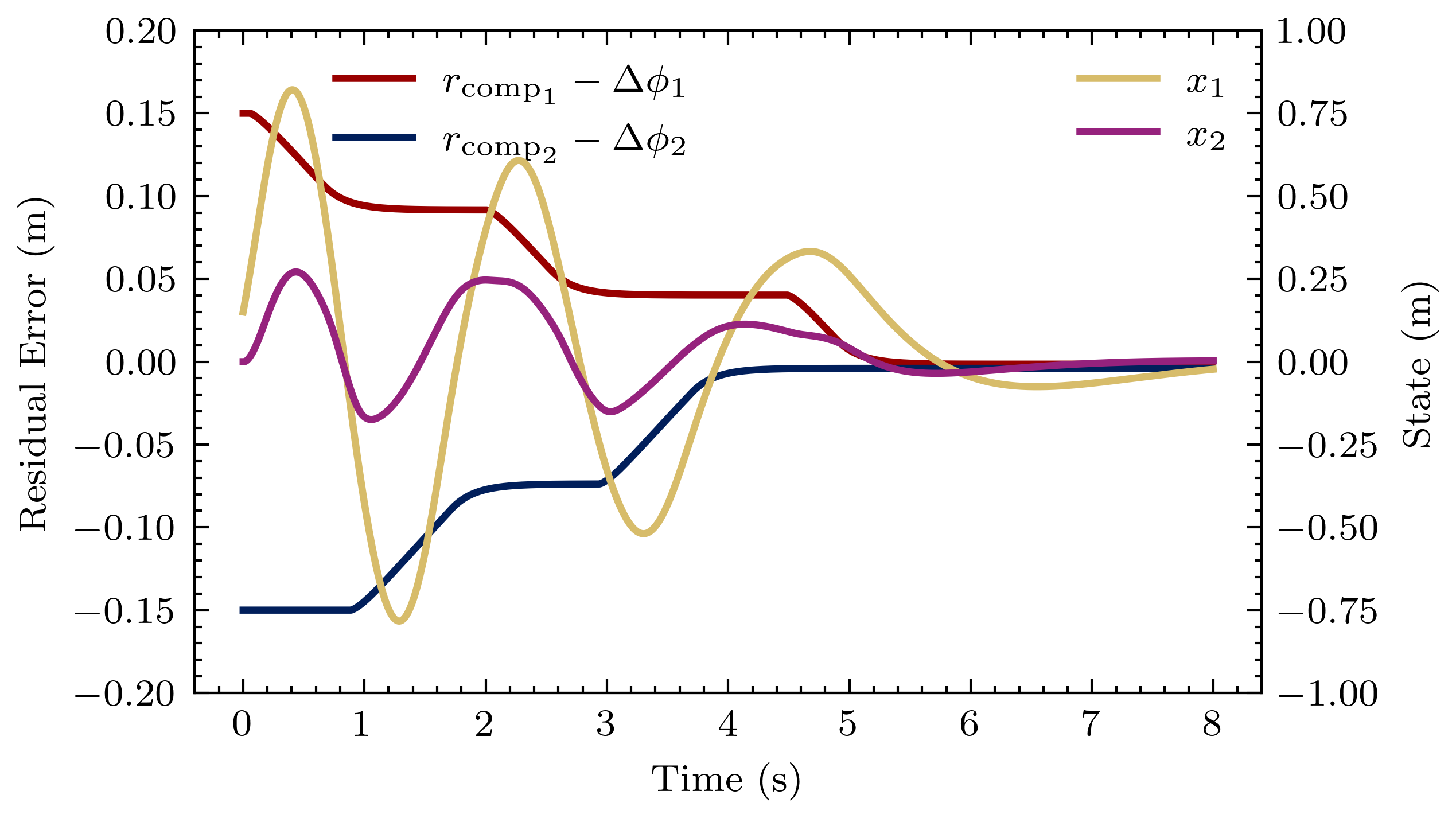}
	\caption{Stabilization of the cart-pole system and convergence of residual. 
 }
	\label{synthetic_convergence}
\end{figure}

We consider a classical cart-pole underactuated system which has been augmented with two soft walls (Figure \ref{grad}). The pole can contact these walls, requiring contact-aware control to stabilize the system (for further details, see \cite{aydinoglu2020stabilization}):
\begin{align*}
  & x_{k+1} = A x_k + B u_k + D \lambda_k + d, \\
  & 0 \leq \lambda_k \perp Ex_k + F \lambda_k + c \geq 0,
\end{align*}
where we have perfect knowledge of the parameters except for $c$. We write $c = \hat c + \Delta \phi$, where $\hat c$ is our initial guess of the parameter.
For the purposes of illustration, we begin with an initial error of $\Delta \phi = [-.15, .15]$.
Notice that the model error that is related to each contact ($\lambda_k$) has a different sign. \rev{This initial model error induces both false positives and false negatives in the model's predictions of contact.} In Figure \ref{grad}, it can be seen that our method successfully returns useful gradient information in all of those scenarios. We highlight that our approach is capable of adapting even when there is no actual contact event (Figure \ref{grad}). We also show that our framework successfully stabilizes the system as well as learning the true residual values (Figure \ref{synthetic_convergence}).



\subsection{Hardware Experiment: Adaptive Trajectory Tracking}
\label{subsec:hw_ball}

Here, we show that our real-time adaptive MPC framework can reliably be used for multi-contact manipulation tasks that require high-speed reasoning about contact events.
Our goal is to roll a rigid ball along a circular trajectory using a Franka Emika Panda Arm (Figure \ref{franka}). \rev{We use PointGrey
cameras to perform vision-based estimation for the ball using Hough transform \cite{duda1972use} and utilize an impedance controller \cite{hogan1985impedance, khatib1987unified} to track high-level commands that our adaptive MPC produces (Figure \ref{C3_high_level_fig_learning}). Experiments are conducted on two desktop computers, one for the adaptive MPC computation and the other for vision tracking and impedance control. For adaptive MPC, we simplify the arm as a point contact. For full details please check Section VII of manuscript \cite{aydinoglu2023consensus}.}

\begin{figure*}[t!]
	\centering
	\includegraphics[width=1.9\columnwidth]{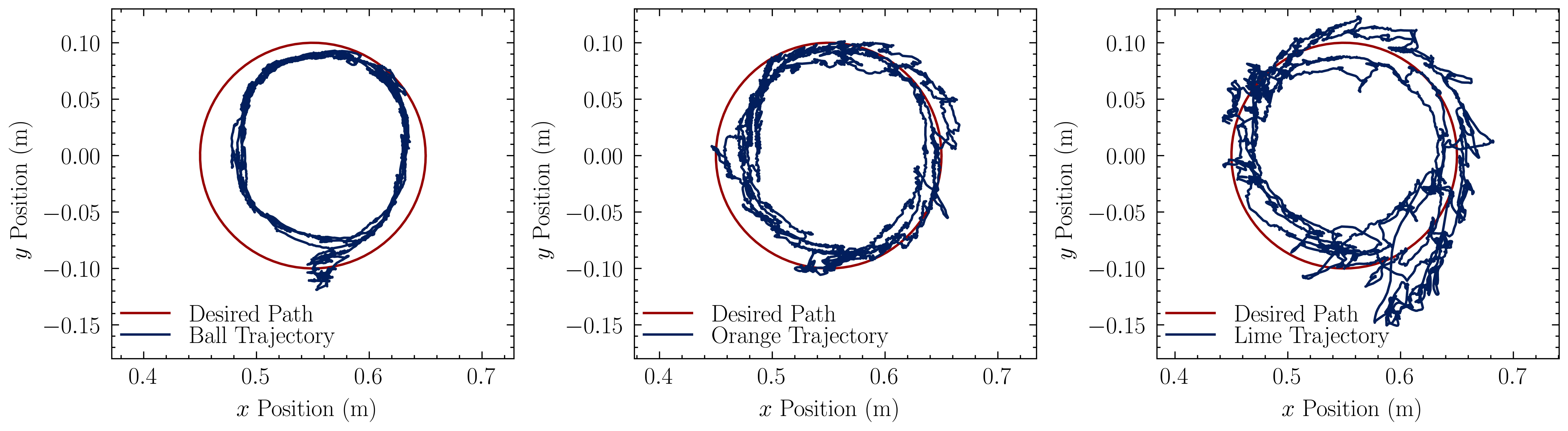}
	\caption{Rolling a rigid ball and fruits (left: ball, middle: orange, right: lime) starting with an inaccurate model. }
	\label{hardware_ball_trajectory}
\end{figure*}

In previous work, careful manual identification of model parameters was required to achieve success \cite{aydinoglu2023consensus}.
For this experiment, the actual radius of the ball is $5$ mm smaller than our parameter estimation. The state estimation for the ball is noisy (vision-based, $80$ Hz), and we do not have an accurate estimation of many model parameters, such as the coefficient of friction.
As a result, \rev{MPC with this incorrect model attempts to push the ball but fails to make contact.} Due to the stiff, hybrid nature of contact dynamics, \rev{MPC is particularly} sensitive to modeling errors that affect the contact/no-contact transition.

\begin{figure}[b!]
 \vspace{-10pt}
	\centering
	\includegraphics[width=0.95\columnwidth]{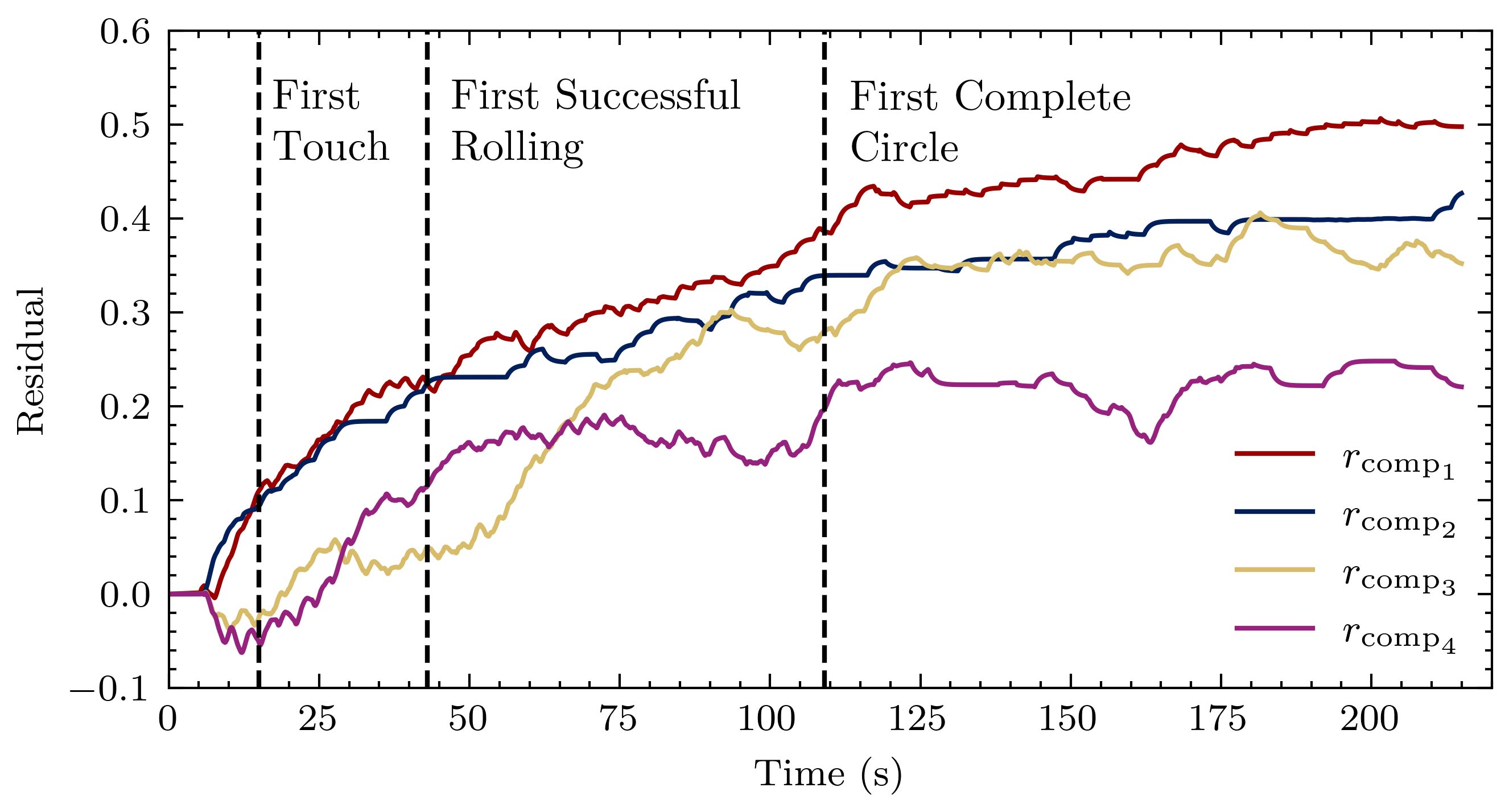}
    \includegraphics[width=0.95\columnwidth]{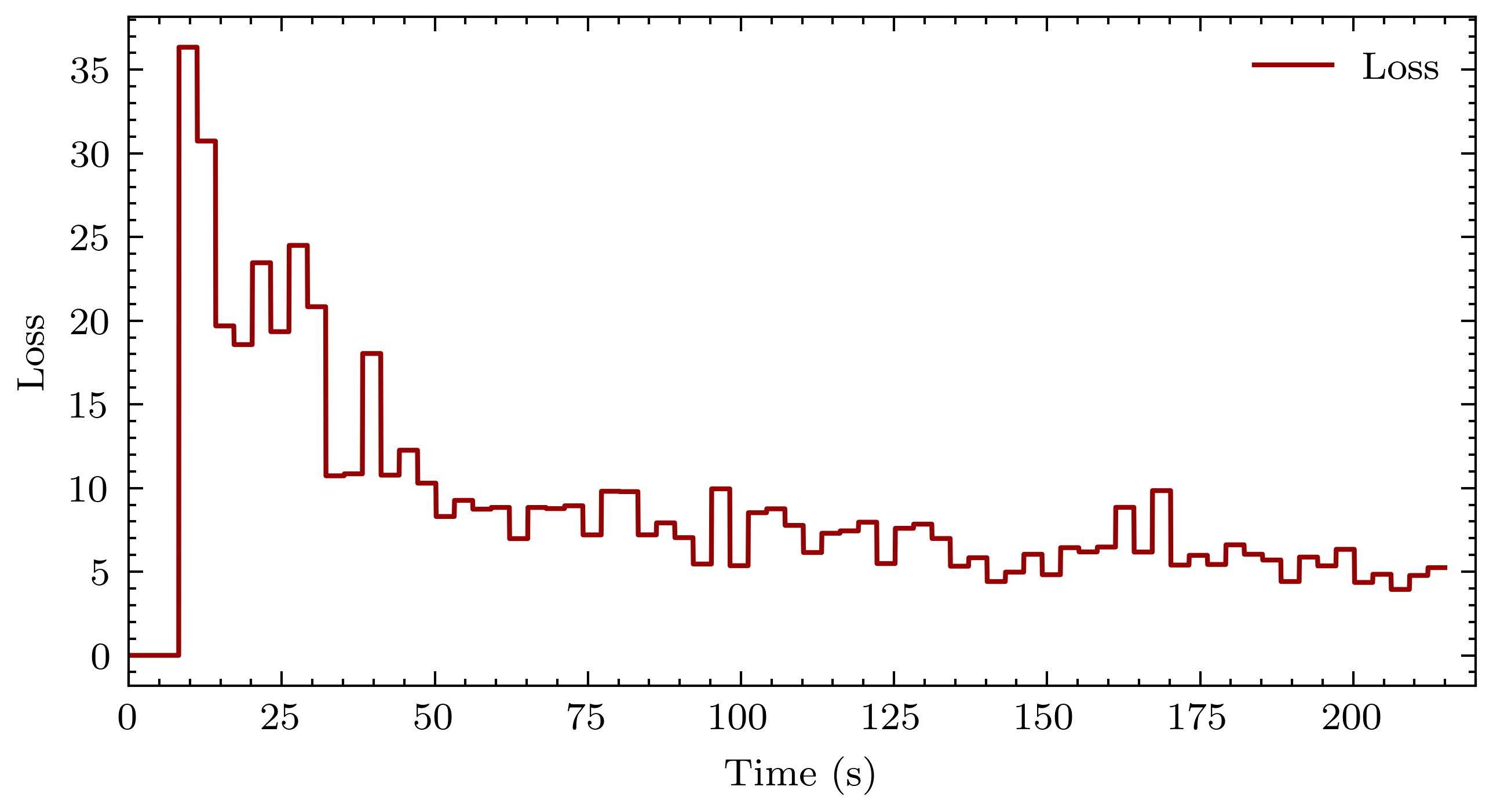}
	\caption{Hardware Experiment: Residual evolution of end-effector and ball contact pair, and loss curve of the learning process. The full experiment lasts for $315$ s. We highlight the adaptation process until loss and residual converge.}
	\label{hardware_ball_residual_loss}
\end{figure}

Without adaptation, model-based control with C3 fails, but our adaptive MPC framework successfully identifies the contact residual and accomplishes the task. The residual learning module (Algorithm \ref{learning_algo}, $20$ Hz), in combination with the C3 module ($80$ Hz), identifies the discrepancy at real-time rates. In the supplementary video, it can be observed that at the beginning the end-effector misses the ball due to inaccurate radius estimation. 
The C3 module then keeps trying to initiate contact. As motivated earlier, the discrepancy between actual motion and predicted motion creates a non-zero gradient and therefore the residual values that correspond to end-effector ball contact pair gradually increase (shown in Figure \ref{hardware_ball_residual_loss}).

In the end, with the learned residual compensating for the inaccurate model, \rev{our approach is able to make contact with the ball, push the ball towards the correct direction to initiate rolling, and roll the ball for $4$ successive circles, successfully accomplishing the task}. Figure \ref{hardware_ball_residual_loss} shows the loss, $\mathcal{L}_\epsilon$, as well as the residual, $r_\text{comp}$, for this experiment. The residual for the end-effector and ball contact shows convergence as the rolling proceeds and the loss curve shows a decreasing trend, proving the effectiveness of our method. We note that in this experiment, the LCP violation rate is $\epsilon = 10^{-7}$, the terms in $Q_d$ related to balls translational velocity are set to $10^{5}$ (others are set to zero), learning rate is $\xi = 10^{-3}$, stiffness parameter is $\gamma = 10^{-2}$ and we have a batch data size of $n_b = 10$. We demonstrate the trajectories of the ball with respect to the desired path in Figure \ref{hardware_ball_trajectory}. \rev{Moreover, with extensive tests, we found that for this experiment, our method with the above-mentioned parameter setting can adapt and compensate well within around 10 pushing trials for model errors up to 8 mm. For larger errors, we either need a more aggressive learning rate or more data (pushing trials) to adapt.}


\subsection{Hardware Experiment: Objects of Non-smooth Surface}

We repeat the previous hardware experiment (Section \ref{subsec:hw_ball}) with slightly deformable objects that have non-smooth surface geometries (Figure \ref{franka}). \rev{Still, our goal is to roll the fruit along the circular path.} We do not have accurate initial estimates of the geometry as our estimate of a fruit's geometry is simply a sphere. For each experiment experiments, we assign an approximate radius for the given fruit and set the sphere radius in our prior model accordingly.


\rev{Due to the bulges and dents on the fruit's surface, the fruit can roll back even after a push in the correct direction. 
We consider successful initiation of rolling to be when the fruit has moved for at least one quarter circle.
MPC without adaptation struggles with the roll back motion, gets stuck in the beginning, and is only able to move the fruit back and forth within a small region near the starting point. In contrast, our method quickly adapts and starts making stronger pushes to roll the fruit (as shown in the supplementary video).} In $90$\% of the $20$ experiments, our method successfully initiated rolling and started tracking the circular path, while MPC without adaptation was never successful ($0$\%).


In addition, fruits tend to produce unpredictable motions at times due to their non-smooth surface geometries, but our method can adapt and accomplish the task. 
Our approach has managed to track at least one circle for $70$\% of the trials with orange and $50$\% of trials with lime ($20$ experiments for each case).
In Figure \ref{hardware_ball_trajectory}, we also report the long-term tracking (4 successive circles) performance of our method with multiple different fruits. The videos of experiments are in the supplementary material.


\section{Conclusion}

We presented an adaptive model predictive control framework for multi-contact systems. The approach uses online residual updates and adaptively compensates model errors and uncertainties at real-time rates. The effectiveness of the method has been shown on multiple hardware experiments, including high-dimensional manipulation problems that include objects with non-smooth surface geometries (fruits).

We have further shown examples where pure model-based control fails but our adaptive strategy leads to success. We have also demonstrated that our learning module is capable of learning the contact model accurately even without actual contact interactions. Also, our parameter estimation can converge to the true parameter values at real-time rates.

We are interested in exploring a wider range of residual models in the future. For example, we might extend our results to include state-dependent terms, or simultaneously learn the non-contact and contact residuals. Integrating our framework with tactile sensors is also an interesting future direction which could speed up our learning process as well as increase the performance of the low-level controller. \cite{aydinoglu2020stabilization}

\bibliographystyle{ieeetr}
\bibliography{bibliography}

\end{document}